\documentclass[conference,a4paper]{IEEEtran}
\IEEEoverridecommandlockouts

\usepackage[hidelinks]{hyperref}
\usepackage[cmex10]{amsmath}
\usepackage{amssymb,amsfonts}
\usepackage{dblfloatfix}

\usepackage[ruled,vlined]{algorithm2e}
\usepackage{graphicx}
\graphicspath{{Figures/PDF/}{Figures/PNG/}}

\usepackage{booktabs}
\usepackage{cite} 
\usepackage{siunitx}
\usepackage[numbers,compress]{natbib}
\usepackage{texnames}
\usepackage{bm,bbm}
\usepackage{orcidlink}
\usepackage{multirow}
\usepackage{tabularx} 
\usepackage{xcolor} 
\usepackage{hyperref} 

\begin{document}

\title{
    {A Vision-Language Framework for \\Multispectral Scene Representation Using\\ Language-Grounded Features}%
}

    

\author{

Enes Karanfil\textsuperscript{1,2} \quad
Nevrez Imamoglu\textsuperscript{1}\orcidlink{0000-0002-2661-599X}
\quad
Erkut Erdem\textsuperscript{2}\orcidlink{0000-0002-6744-8614}
\quad
Aykut Erdem\textsuperscript{3}\orcidlink{0000-0002-6280-8422}\\
\texttt{\small \{enes.karanfil,nevrez.imamoglu\}@aist.go.jp} \quad
\texttt{\small erkut@cs.hacettepe.edu.tr} \quad
\texttt{\small aerdem@ku.edu.tr}\\

$^1$AIST, Tokyo, Japan \quad
$^2$Hacettepe University, Ankara, Turkey \quad
$^3$Ko\c{c} University, Istanbul, Turkey
}

\maketitle
\begin{abstract}
Scene understanding in remote sensing often faces challenges in generating accurate representations for complex environments such as various land use areas or coastal regions, which may also include snow, clouds or haze. To address this, we present a vision-language framework named Spectral-LLaVA, which integrates multispectral data with vision-language alignment techniques to enhance scene representation and description. Using the BigEarthNet-v2 dataset from Sentinel-2, we establish a baseline with RGB-based scene descriptions and further demonstrate substantial improvements through the incorporation of multispectral information. Our framework optimizes a lightweight linear projection layer for alignment while keeping the vision backbone of SpectralGPT frozen. Our experiments encompass (1) scene classification using linear probing, and (2) language modeling for jointly performing scene classification and description generation. Our results highlight Spectral-LLaVA's ability to produce detailed and accurate descriptions, particularly for scenarios where RGB data alone proves inadequate, while also enhancing classification performance by refining SpectralGPT features into semantically meaningful representations. The code and dataset for this project are available 
\href{https://enkaranfiles.github.io/Spectral_LLaVA/}{\textcolor{magenta}{here}}.


\end{abstract}

\section{Introduction}
Remote sensing has become indispensable in numerous real-world applications, including agriculture~\cite{peartrees}, urban planning, environmental monitoring~\cite{deforestation}, and disaster management~\cite{sen1}, leveraging satellite imagery captured by platforms such as Landsat, Sentinel-1, Sentinel-2, and ALOS. Recent advances in multi-modal large language models, including the LLaVA family~\cite{liu2023improvedllava, liu2024llavanext, liu2023visualinstruction} and BLIP-3 \cite{blip3}, have opened new opportunities for applying vision-language models to remote sensing tasks. Studies such as  \cite{kuckreja2023geochat,elgendy2024geollava,luo2024skysensegpt} illustrate the potential of these models for effective scene description. However, most vision-language models in remote sensing rely predominantly on RGB imagery, overlooking the rich multispectral data readily available. While RGB-based models have proven effective in tasks like visual question answering, scene description generation, and classification, multispectral data offers complementary and crucial information for earth observation, capturing spectral bands beyond the visible spectrum. Utilizing this data effectively remains a challenge due to its complexity, requiring significant domain expertise and manual effort. This challenge becomes even more pronounced in tasks such as scene classification and description generation in multispectral domains, which demand models capable of leveraging the inherent spectral diversity of such data.


\begin{figure}[t]
	\centering
	\includegraphics[width = 0.85\columnwidth, trim=0 0 0 0, clip]{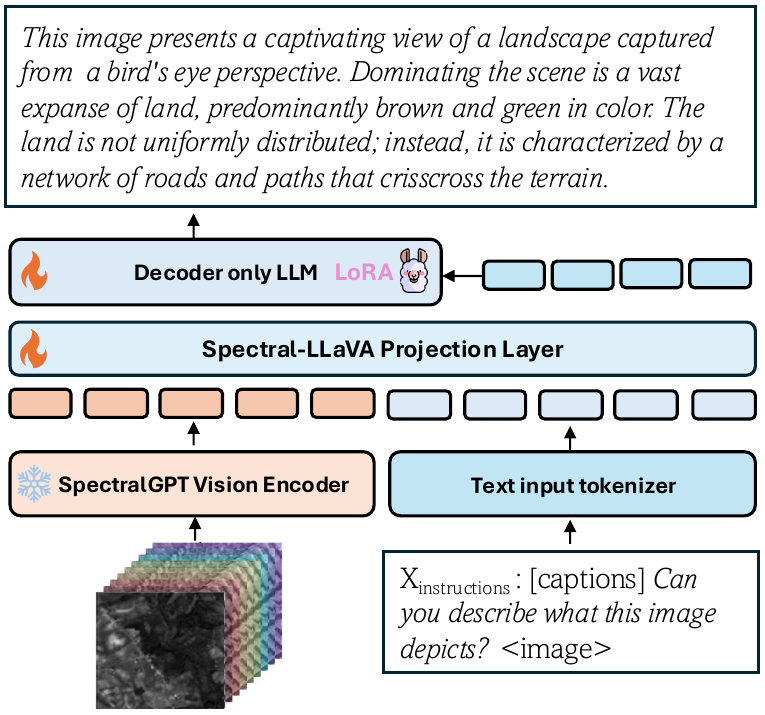}
	\centering
    \caption{Proposed Spectral-LlaVA vision-language framework for multispectral images.}\label{fig:method}
\end{figure}

To address this gap, we propose Spectral-LLaVA, a framework that extends the popular LLaVA~\cite{liu2023visualinstruction} model to the multispectral domain (Figure~\ref{fig:method}). By leveraging the BigEarthNet dataset~\cite{clasen2021reben}, we constructed a novel multispectral scene description dataset. Detailed scene descriptions were generated using the ShareCaptioner-ShareGPT4V model~\cite{chen2023sharegpt4v}, conditioned on metadata and RGB channels. These descriptions were then aligned with visual features via a vision-language alignment process to create language-grounded features. To evaluate the utility of this dataset and framework, we conducted classification experiments using spectral domain features before and after alignment. The results demonstrate that language-grounded features significantly enhance semantic richness. Additionally, by fine-tuning a decoder-only large language model, we developed a vision-language model capable of robust classification and scene description in the multispectral domain. Notably, this was achieved without updating the vision encoder, showcasing the framework's adaptability to multispectral data with minimal changes.

To summarize, the key contributions of our work are as follows:

\begin{itemize}
    \item We generate a novel instruction-tuning dataset aimed at equipping vision language models with the ability of understanding multispectral data.
    \item We introduce a first-of-its-kind spectral-domain vision-language framework designed to effectively process multispectral data.
    \item We demonstrate the benefits of language-grounded features, which lead to improved performance in classification and scene description generation tasks on multispectral imagery.
\end{itemize}


\section{Related Work}

\subsection{Remote Sensing Foundation Models}

Recent advancements in vision foundation models have significantly enhanced remote sensing by addressing spectral complexity and large-scale data challenges. For instance, the Convolutional Transformer Joint Network (CTJN)~\cite{ctjn2022} improves spectral reconstruction by integrating convolutional and transformer modules for spatial-spectral coupling. RingMo~\cite{sun2022ringmo} adapts generative self-supervised learning to effectively handle dense and small objects in remote sensing imagery, achieving state-of-the-art performance. Masked Autoencoders (MAE)~\cite{he2021mae} employ masked image modeling for scalable self-supervised learning, demonstrating strong generalization capabilities even on multispectral remote sensing data as in Prithvi-EO-2.0~\cite{szwarcman2024prithvieo} and SpectralGPT~\cite{spectralgpt}. Prithvi-EO-2.0~\cite{szwarcman2024prithvieo} integrates temporal and spatial embeddings for multi-temporal Earth observation, excelling across various geospatial tasks. Among these, SpectralGPT~\cite{spectralgpt} stands out for its spectral representation power, utilizing a 3D generative Transformer to leverage spatial-spectral coupling and progressive training. This design facilitates downstream tasks such as classification, segmentation, and change detection. We selected SpectralGPT as our vision backbone as a result of its robust spectral modeling within the MAE framework.

\subsection{Vision-Language Models for Remote Sensing}
The development of vision-language models and multimodal large language models has advanced remote sensing image analysis, enabling tasks such as image captioning, VQA, visual grounding, and change detection. Recent efforts focus on bridging the domain gap between natural and remote sensing imagery through comprehensive datasets. For example, RSGPT~ \cite{hu2023rsgpt} and EarthMarker \cite{zhang2024earthmarker} highlight high-quality datasets like RSICap, which comprises human-annotated captions, and RSVP, which offers multi-granularity interpretation. These resources enhance spatial and semantic reasoning for tasks such as scene graph generation and visual prompting.

Other work emphasizes task-specific capabilities and modality alignment. Models like RS-MoE \cite{lin2024rsmoe}, SkySenseGPT \cite{luo2024skysensegpt}, and LHRS-Bot-Nova \cite{li2024lhrsbotnova} improve feature extraction and fine-grained reasoning for tasks like remote sensing captioning by employing instruction-tuning. and modular architectures. GeoGround \cite{zhou2024geoground}, GeoChat \cite{kuckreja2023geochat}, and SkyEyeGPT \cite{zhan2024skyeyegpt} address multitask challenges by extending vision-language models for region-level queries, utilizing formats such as bounding boxes and segmentation masks. Temporal reasoning models, including  RingMoGPT \cite{wang2023ringmogpt}, TEOChat \cite{irvin2024teochat}, and GeoLLaVA \cite{elgendy2024geollava} excel in tasks like change detection and temporal classification.

Collectively, these models represent substantial advancements in tackling remote sensing-specific challenges by leveraging sophisticated architectures, curated datasets, and enhanced visual-language alignment, enabling effective performance across diverse spatial, semantic, and temporal tasks.

\section{Method of \textit{Spectral-LLaVA}}

\subsection{\textit{Spectral-LLaVA} Architecture}

\subsubsection{Visual Backbone} 

The visual backbone of Spectral-LLaVA leverages the encoder component of SpectralGPT \cite{spectralgpt} to extract multispectral features. This encoder is designed to learn robust spectrally-aware visual representations from multispectral data, capturing essential spectral and spatial correlations. Unlike the original SpectralGPT framework, which includes masking and reconstruction, Spectral-LLaVA focuses solely on the encoder's pre-trained representations for feature extraction.

\subsubsection{Multimodal Projector}

Following the LLaVA \cite{liu2023visualinstruction} framework, Spectral-LLaVA employs a trainable linear projection layer to align visual and language modalities. For an input image $X_v$, the pre-trained SpectralGPT encoder extracts multispectral features $Z_v = g(X_v)$. Then, a projection matrix $W$ is  applied to convert $Z_v$ into language embedding tokens $H_v$, matching the dimensionality of the word embedding space in the language model \cite{liu2023visualinstruction}. 

\subsubsection{Large Language Model (LLM)}

Spectral-LLaVA utilizes the LLaMA3\cite{grattafiori2024llama3} model as its language backbone. This decoder-only large language model is fine-tuned to integrate multispectral features and perform downstream tasks.

\subsection{Training Recipe}

We utilize the SpectralGPT vision encoder, pretrained for 200 epochs on fMoW \cite{fmow2018} using the MAE architecture, followed by continual pretraining for an additional 100 epochs on BigEarthNet-v1\cite{sumbul2019bigearthnet}. Fine-tuning is performed with LLaMA 3 as the language model, employing LoRA-based parameter adaptation with a rank r as 128.

For the alignment stage, an effective batch size of 8 is used, along with a linear projection layer based on the LLaVA architecture. In the conversational fine-tuning stage, image samples are resized to 128 × 128, and the model is trained for 1 epoch with an effective batch size of 64. The training process is optimized using the Adam optimizer.

\section{Generating Spectral Domain Multimodal-Instruction Dataset}

Recent advances in remote sensing and multimodal learning have seen a growing interest in large language models (LLMs) tuned for specific tasks using customized instruction datasets. While several studies have developed instruction-tuning datasets \cite{zhan2024skyeyegpt, hu2023rsgpt, elgendy2024geollava, kuckreja2023geochat} tailored to remote sensing applications, most of these efforts have predominantly focused on optical imagery. Notably, the EarthGPT \cite{zhang2024earthgpt} study introduced MMRS-1M, a multi-modal, multi-sensor instruction-following dataset with over 1 million image-text pairs spanning optical, SAR, and infrared imagery, advancing multimodal remote sensing research. However, a significant gap remains in the availability of vision-language datasets explicitly designed for multispectral imagery, which is critical for applications that require detailed spectral analysis.

To address this limitation, we present the Spectral-Inst dataset, a multimodal instruction dataset that integrates multispectral satellite imagery with detailed textual annotations. By focusing on multispectral data, Spectral-Inst aims to bridge the gap between vision-language models and the unique demands of remote sensing tasks, enabling advancements in areas such as environmental monitoring, land-use classification, and spectral feature interpretation. To create the Spectral-Inst dataset, we built upon the robust foundation provided by the BigEarthNet-v2 \cite{clasen2021reben} dataset. With its 549,488 multispectral satellite images captured by Sentinel-2.

The pipeline starts by converting multispectral images into RGB-domain optical images, a critical step to standardize the input format and ensure compatibility with the state-of-the-art image captioning model ShareCaptioner, part of the ShareGPT4V project \cite{chen2023sharegpt4v}. This model is then employed to generate detailed captions that describe the scene content. To improve caption accuracy and semantic richness, we integrate metadata—including image labels and spatial attributes—into the captioning process. A model-generated language dataset, considered pseudo-data due to its uncertain accuracy, demonstrates utility through experiments with language-grounded features. Results show that integrating generated captions enhances visual features semantic representation, highlighting their value in contextual understanding.


\section{Experiments}

\subsection{Instruction Tuning}

\subsubsection{Image-Text Alignment}
The alignment stage in Spectral-LLaVA pairs images with language instructions, prompting descriptions guided by corresponding captions as ground truth. During this process, the visual encoder and LLM weights remain frozen, and only the projection matrix is trained to align image features with the LLM's word embedding space. This results in language-grounded visual features, enabling effective interaction between visual and textual modalities.\\

For the purposes of the ablation study, two separate projector modules were trained.

\begin{itemize}
    \item \textbf{Language-Grounded Features Derived from Class Labels}: The alignment layer generates language-grounded features by training with multi-label class annotations from the BigEarthNet-v2 dataset.
    \item \textbf{Language-Grounded Features Derived from Scene Descriptions}: The alignment layer utilizes scene descriptions as textual inputs to map image features into the LLM's word embedding space, fostering semantically enriched and contextually grounded representations. 
\end{itemize}

\begin{figure}[t!]
	\centering
	\includegraphics[width = 0.5\textwidth, trim=0 0 0 0, clip]{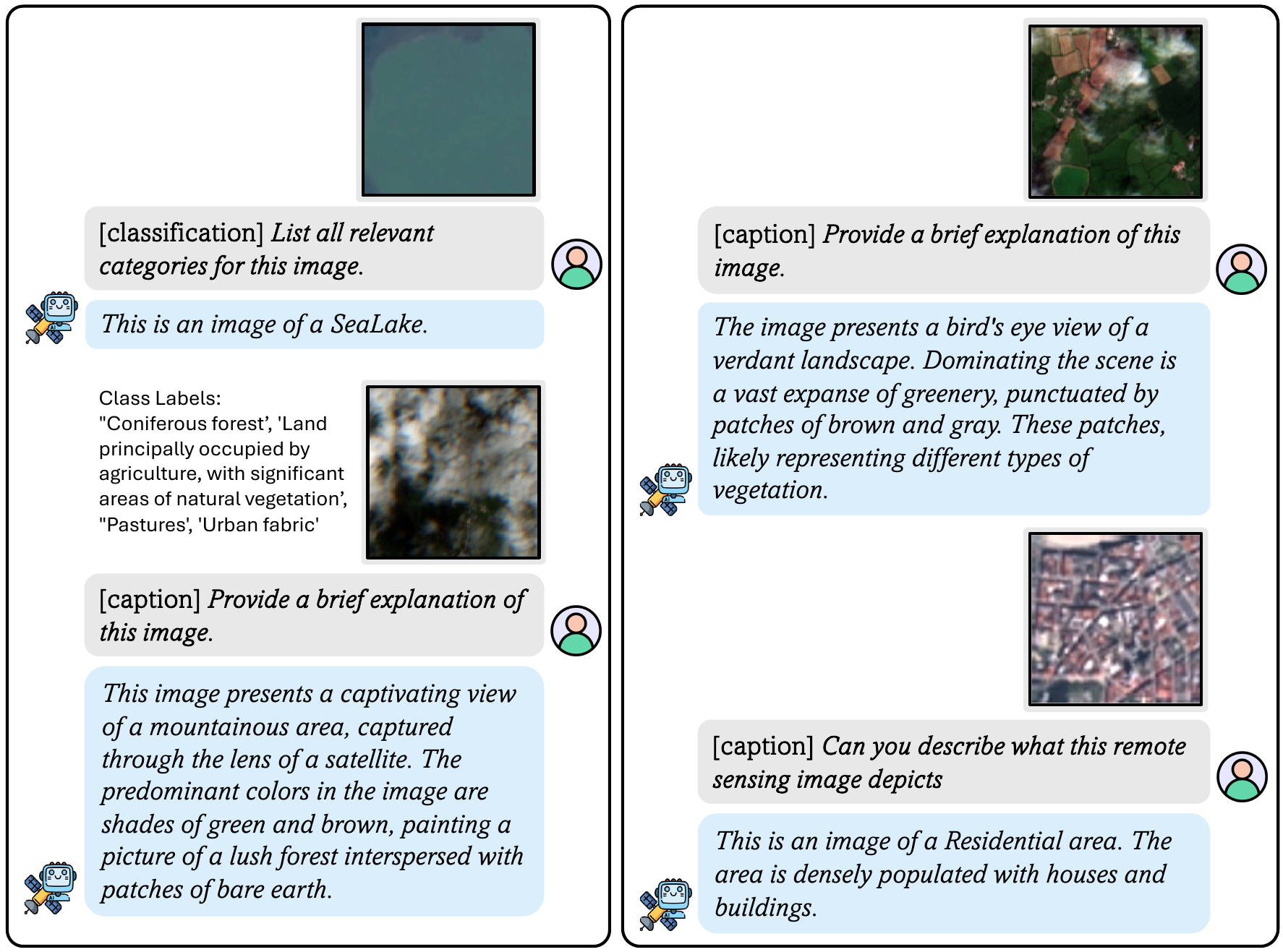}
	\centering
    \caption{Qualitative chat samples with given multispectral images (images in the figure are contrast enhanced RGB image version of multispectral data just for visualization.). Sample descriptions are given by taking the first a few sentences of model output just to provide visual examples.}\label{fig:qualitative_samples}
\end{figure}

\begin{table*}[t!]
    \centering
    \caption{Comparison of original and revised LLaVA-Bench detail scores with alignment stages.}
    \renewcommand{\arraystretch}{1.3} 
    \setlength{\tabcolsep}{5pt} 
    \resizebox{0.98\textwidth}{!}{ 
    \begin{tabular}{llccl} 
        \hline
        \textbf{Method} & \textbf{Domain} & \textbf{Original} & \textbf{Revised} & \textbf{Description} \\ \hline
        \textbf{GeoChat}\cite{kuckreja2023geochat} & RGB & 42.0 & 27.5 & LLaVA-based vision-language model for remote sensing optical domain. \\ \hline
        \textbf{Spectral-LLaVA$_{\text{ClassLabel}}$} & Multispectral & 64.0 & 55.4 & Alignment stage done with class labels textual descriptions only. \\ 
        \textbf{Spectral-LLaVA$_{\text{SceneDesc}}$} & Multispectral & \textbf{72.4} & \textbf{61.6} & Alignment stage done with scene descriptions by ShareCaptioner. \\ \hline
    \end{tabular}}
    \label{tab:llava_bench}
\end{table*}

\subsubsection{Multimodal Finetuning}
In instruction tuning stage, the Spectral-LLaVA model undergoes end-to-end fine-tuning to enhance its multimodal reasoning capabilities for specific tasks, such as scene description and scene classification. During this stage, the visual encoder remains frozen, while the projection layer and LLM are updated to align with task-specific objectives. Training incorporates task tokens, such as \textbf{[caption]} and \textbf{[classification]}, not as traditional classification tokens but as contextual indicators, allowing the language model to establish relationships between tasks.

\textbf{Results:} We present quantitative results in Table 1, derived from the subset of the BigEarthNet-v2 test set, alongside qualitative examples in Figure 2 showing model outputs. Since ground truth explanation data is unavailable for this domain, we evaluated the generated descriptions by examining how well they aligned with the class labels in the dataset. The evaluation focused on criteria such as helpfulness, relevance, accuracy, and level of detail, providing measure of the model's performance.
Quantitative results in Table \ref{tab:llava_bench} show that the Spectral-LLaVA model demonstrated better performance in the benchmark tested with GeoChat using RGB image version of multispectral data. Additionally, the quality of the produced captions improved when scene descriptions was utilized in the projection layer, highlighting its effectiveness in generating detailed and relevant explanations. Class labels proved more precise for initial semantic alignment but lacked the contexual richness necessary for generative tasks after fine-tuning, highlighting the importance of richer input representations during these stages.

\begin{figure}[t]
	\centering
	\includegraphics[width=0.95\columnwidth, trim=0 0 0 0, clip]{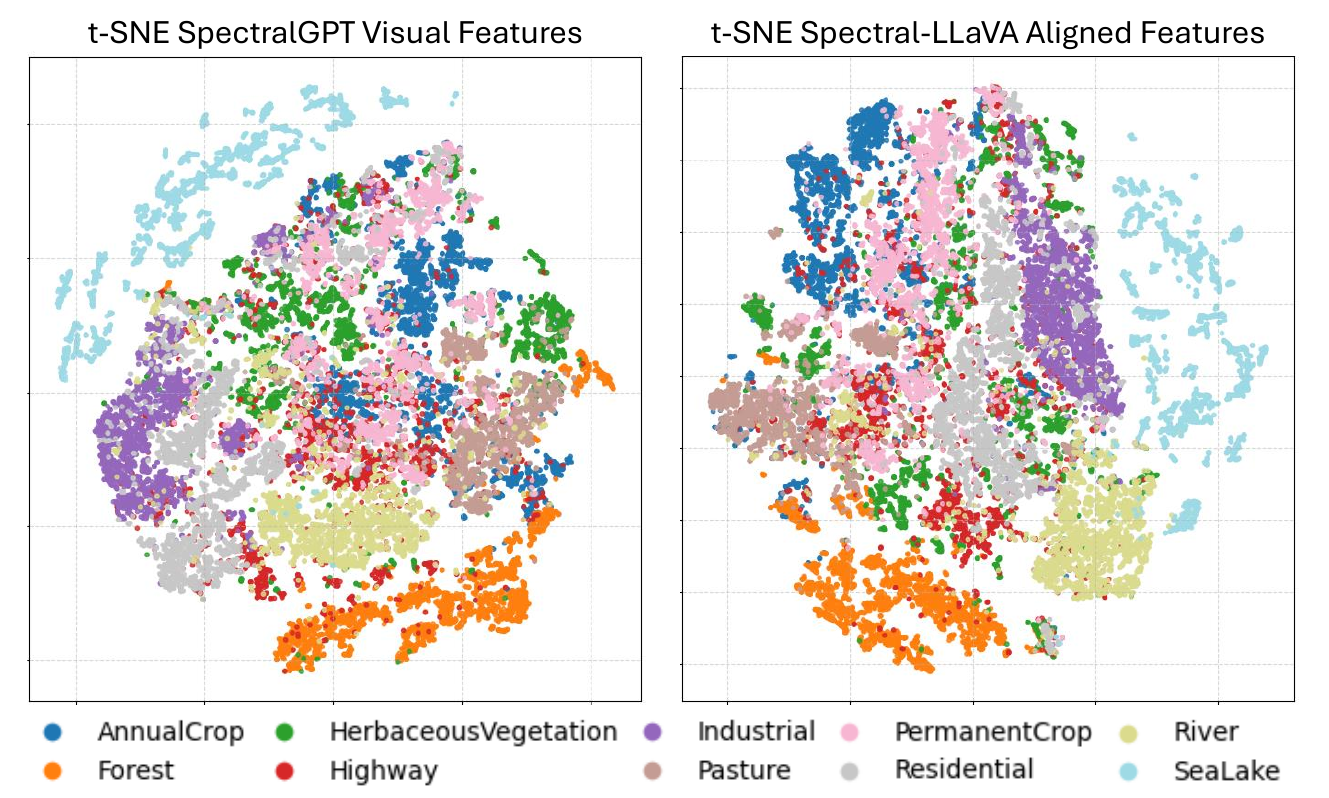}
	\caption{Comparison of visual features and aligned features projected in a 2D space using t-SNE for EuroSAT miltispectral data. The alignment highlights the transformation of raw visual features into a domain-aligned latent space, showcasing clustering improvements.}
	\label{fig:tSNE}
\end{figure}

\subsection{Scene Classification}
In addition to the finetuned model evaluation, we also compare the representation ability of Spectral-LLaVA language-grounded features and visual only features (i.e. SpectralGPT \cite{spectralgpt} encoder output) to verify the quality of representation ability on classification task with EuroSAT dataset \cite{helber2019eurosat}. This dataset consists of 27000 multispectral images where each image is labeled with one of the 10 semantic categories of land use and land cover type as the details are given in \cite{helber2019eurosat}. 

\textbf{Results:} First, as a qualitative evaluation, we employ t-SNE\footnote{\url{https://scikit-learn.org/stable/modules/generated/sklearn.manifold.TSNE}} on SpectralGPT vision only features and Spectral-LLaVA language-grounded features derived from class labels for EuroSAT data samples with given category labels. Figure \ref{fig:tSNE} shows better clustered categorial structure on t-SNE distribution of language-grounded featues compared to vision-only SpectralGPT encoder features. 

\begin{figure}[t!]
	\centering
	\includegraphics[width = \columnwidth, trim=25 10 0 0, clip]{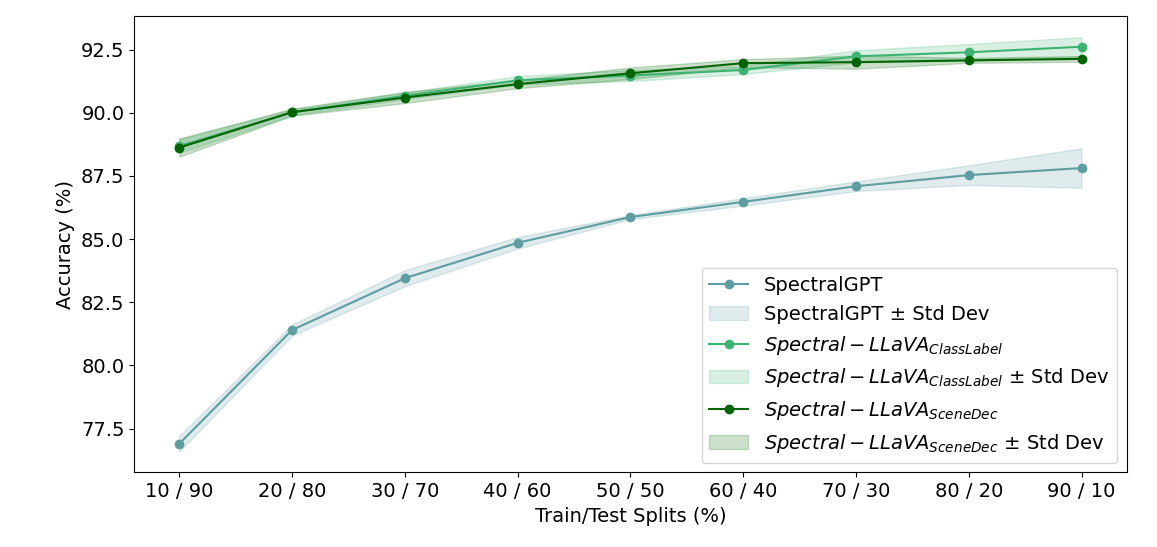}
	\centering
    \caption{Test Accuracy Comparison for SpectralGPT and Spectral-LLaVA features through linear probing on EuroSAT dataset with 5-fold cross-validation using various train-test split ratios.}\label{fig:eurosat_comparison}
\end{figure}

In addition, to check representation ability quantitatively, simple linear probing (a linear neural network layer) is applied on features of vision encoder or language-grounded features to classify given features. Linear layer parameters are trained with Adam optimizer\footnote{\url{https://pytorch.org/docs/stable/generated/torch.optim.Adam.html}} with 100 batch size with a learning rate of 0.0001 with 100 epochs. We evaluate the classification accuracy with 5-fold cross-validation on various train-test split ratios starting from the minimum value of 10\% as train sample size to maximum train split ratio of 90\% by 10\% increments. Fig. \ref{fig:eurosat_comparison} demonstrates that EuroSAT test data classification performance of language-grounded features (both pretrained based on class label annotations and scene descriptions) outperforms vision only SpectralGPT features with a significant margin at every train-test split ratios. This highlights the fact that  image to language alignment provides or adds some useful semantic information on top of the vision only features, which is correlated with the multispectral image content.

\section{Conclusion}
The rapid progress in vision-language has profoundly impacted remote sensing, driving advancements in downstream tasks such as VQA, classification, change detection, and object detection within the optical RGB image domain. Despite these strides, the application of vision-language models to multispectral imagery remains largely unexplored. To the best of our knowledge, proposed Spectral-LLaVA framework represents the first vision-language model explicitly developed for the spectral domain in remote sensing. We curate and share Spectral-Inst, an instruction-tuning dataset tailored for multispectral data understanding. Additionally, we demonstrate that language-grounded features can effectively be utilized for classification tasks, showcasing how the Spectral-Inst dataset produced successfully maps semantic visual features into a meaningful representational space.

\section{Acknowledgement}
This work supported by AIST policy-based budget project ``R\&D on Generative AI Foundation Models for the Physical Domain". 

\bibliographystyle{IEEEtranN}
\bibliography{references}
\end{document}